# Der Effizienz- und Intelligenzbegriff in der Lexikographie und künstlichen Intelligenz: kann ChatGPT die lexikographische Textsorte nachbilden?

Iván Arias-Arias, *Instituto da Lingua Galega, Universidade de Santiago de Compostela, Spanien (ivanarias.arias@usc.gal)*
*(https://orcid.org/0000-0003-2673-0899)*

María José Domínguez Vázquez, *Instituto da Lingua Galega, Universidade de Santiago de Compostela, Spanien (majo.dominguez@usc.es)*
*(https://orcid.org/0000-0002-6060-9577)*
und
Carlos Valcárcel Riveiro, *Universidade de Vigo, Spanien*
*(carlos.valcarcel@uvigo.gal) (https://orcid.org/0000-0003-1123-5211)*

**Zusammenfassung:** Mittels Pilotexperimente für das Sprachenpaar Deutsch–Galicisch untersucht der vorliegende Aufsatz den Effizienz- und Intelligenzbegriff in der Lexikographie und künstlichen Intelligenz (KI). Die Experimente versuchen, empirisch und statistisch fundierte Erkenntnisse über die lexikographische Textsorte „Wörterbuchartikel" in den Antworten von ChatGPT-3.5 zu gewinnen, und darüber hinaus über die lexikographischen Daten, mit denen dieser Chatbot trainiert wurde. Zu diesem Zweck werden sowohl quantitative als auch qualitative Methoden herangezogen. Der Analyse liegt die Auswertung der Outputs von mehreren Sessions mit demselben Prompt in ChatGPT-3.5 zugrunde. Zum einen wird die algorithmische Leistung von intelligenten Systemen im Vergleich zu Daten aus lexikographischen Werken bewertet; zum anderen werden die gelieferten ChatGPT-Daten über konkrete Textteile der genannten lexikographischen Textsorte analysiert. Die Resultate dieser Studie tragen dazu bei, nicht nur den Effizienzgrad von diesem Chatbot hinsichtlich der Erstellung von Wörterbuchartikeln zu evaluieren, sondern auch in den Intelligenzbegriff, die Denkprozesse und die in beiden Disziplinen auszuführenden Handlungen zu vertiefen.

**Stichwörter:** LEXIKOGRAPHIE, KI, CHATGPT-3.5, WÖRTERBUCHARTIKEL, EFFIZIENZBEGRIFF, INTELLIGENZBEGRIFF, LEXIKOGRAPHISCHE TEXTSORTE, TRAININGSDATEN, LEXIKOGRAPHISCHE DATEN

**Abstract: Efficiency and Intelligence in Lexicography and Artificial Intelligence: Can ChatGPT Recreate the Lexicographical Text Type?** By means of pilot experiments for the language pair German–Galician, this paper examines the concept of efficiency and intelligence in lexicography and artificial intelligence (AI). The aim of the experiments is to gain empirically and statistically based insights into the lexicographical text type "dictionary article" in the responses of ChatGPT-3.5, as well as into the lexicographical data on which this chatbot was trained. Both quantitative and qualitative methods are used for this purpose. The analysis is based on the evaluation of the outputs of several sessions with the same prompt in ChatGPT-3.5. On the one hand, the algorithmic performance of intelligent systems is evaluated in comparison with data from lexicographical works; on the other hand, the ChatGPT data supplied is analysed using specific text passages of the aforementioned lexicographical text type. The results of this study not only help to evaluate the efficiency of this chatbot regarding the creation of dictionary articles, but also to delve deeper into the concept of intelligence, the thought processes and the actions to be carried out in both disciplines.

## 1. Einführung

In den letzten Jahrzehnten hat sich die Künstliche Intelligenz (KI) explosionsartig entwickelt, so dass sogar einige mit ihr verbundenen Entwicklungen Bestandteil unseres alltäglichen Lebens geworden sind. Somit kennen oder verwenden wir alle z.B. digitale Assistenten wie *Alexa*, *Siri* oder *Google Assistant*. Der Einfluss der KI auf unseren Alltag ist unbestritten, auch wenn wir uns dessen häufig nicht bewusst sind; ihr Einfluss auf die lexikographische Tätigkeit hat eine rege Diskussion ausgelöst. Manche Autoren sind der Ansicht, dass der Output von KI-Tools wie dem Chatbot ChatGPT-3.5[1] (*Generative Pretrained Transformer*, https://chat.openai.com/) statistisch gesehen die Ergebnisse einiger Wörterbücher übertrifft (vgl. Phoodai und Rikk 2023) und manche prophezeien sogar das Ende der Lexikographie (vgl. de Schryver und Joffe 2023).

    In diesem Zusammenhang geht es hier um die Frage der Effizienz (s. 3) von ChatGPT bei einem konkreten Prompt[2] (Eingabeaufforderung) betreffend den Wörterbuchartikel als lexikographische Textsorte (vgl. Wiegand 1996). Konkret fragen wir, inwiefern die ChatGPT-Ergebnisse für das Deutsche und das Galicische mit der Mikrostruktur von Referenzwörterbüchern (DUDEN-Onlinewörterbuch, https://www.duden.de/woerterbuch, fortan DUDEN und Dicionario da Real Academia Galega, https://academia.gal/dicionario/rag, fortan DRAG) quantitativ bzw. qualitativ vergleichbar sind. Eine Analyse der lexikographischen Daten sowie ihr Vergleich mit denen in Referenzwörterbüchern ist unseres Erachtens die erste Frage, die überhaupt gestellt werden sollte. Daher stehen die Datenverfügbarkeit sowie ihre Gegenüberstellung im

Mittelpunkt, jedoch nicht die Analyse der möglichen Verwendung von ChatGPT bei lexikographischen Aufgaben.

Dazu gliedert sich der Aufsatz wie folgt: in Kapitel 2 wird ein Gesamtüberblick über ausgewählte Schwerpunkte der Lexikographie und der KI dargestellt. Dem Effizienz- und der Intelligenzbegriff widmet sich einführend Kapitel 3: die Frage lautet, ob ChatGPT über Daten bezüglich der Textsorte „Wörterbuchartikel" verfügt. Kapitel 4 dient der Erklärung der Methode und bietet die statistisch ausgewertete Untersuchung der Ergebnisse aus dem Chatbot an. Die Hauptergebnisse der quantitativ und qualitativ ausgerichteten Analyse werden in Kapitel 5 vorgestellt. Schlussfolgerungen werden in Kapitel 6 gezogen.

## 2.    Zur Lexikographie und zur künstlichen Intelligenz: Einführendes

Das WLWF-3 (2020: 224) definiert Lexikographie als „Menge aller Aktivitäten, die auf die Erstellung lexikographischer Nachschlagewerke gerichtet ist". Neben der theoretischen Untersuchung auf Gebieten wie der Metalexikographie, Wörterbuchforschung, Wörterbuchkritik u.a. und dem Erwägen und der Anwendung verschiedenartiger Strategien, Methoden und Techniken zur Analyse des Sprachmaterials befasst sich die lexikographische Tätigkeit mit der Entwicklung von Informationssystemen (vgl. Villa Vigoni 2018, https://www.emlex.phil.fau.de/ueberuns/publikationen/andere-publikationen/), in jeglichem Format und für verschiedenartige Zugangsgeräte (Wörterbücher, Portale, Wörterbuch-Apps u.a.), bei denen i.d.R. menschliche Benutzende Antwort auf eine Suchanfrage finden können (s. 5). Sowohl die adäquate Datenpräsentation angesichts der Wörterbuchziele und -adressaten, als auch die Datenqualität bzw. Verantwortung für ihre Qualität (Kouassi 2022) stehen als lexikographische Aufgaben im Mittelpunkt. Die Lexikographie wird zudem als eine wissenschaftliche und kulturelle Praxis aufgefasst, die die Entstehung der Werke sowie ihren Gebrauch ermöglichen sollte (Wiegand 1983: 38), und dabei trägt sie auch eine gesellschaftliche Verantwortung. Die Rolle des Wörterbuchs als Autorität ist in diesem Zusammenhang nicht zu übersehen (Kosem et al. 2019).

Hinsichtlich des anvisierten Benutzerkreises lassen sich die Endprodukte der Lexikographie in Bezug auf Sprachniveaus, Ziele u.a. klassifizieren, aber auch bezüglich der Unterscheidung zwischen Computerlexikographie und computergestützter Lexikographie. In diesem Zusammenhang kann man behaupten, dass Wörterbücher für menschliche BenutzerInnen die primären Endprodukte der computergestützten Lexikographie sind (s. 5). Hingegen befasst sich die Computerlexikographie mit der Entwicklung von Ressourcen für die maschinelle Weiterverarbeitung. Solche maschinenlesbaren Lexika[3] finden Anwendung in verschiedenen NLP-Aufgaben wie *Parsing*, maschineller Übersetzung oder Sprachgenerierung. Folglich stellen sich Maschinen als primäre

Adressaten (nicht BenutzerInnen) der Ergebnisse der Computerlexikographie heraus (Weiteres dazu in 5).

Daraus folgt, dass die Zusammenstellung von lexikographisch akkuraten Daten zwecks ihrer maschinellen Lesbarkeit auch als eine mögliche Aufgabe der Lexikographie hervortritt. Eigentlich ist das nicht neu, denn schon Mel'čuk (1984) hat sein *Dictionnaire explicatif et combinatoire du français contemporain* mit Blick auf diese mögliche Anwendung der lexikographischen Daten entwickelt.

Im Gegensatz zur Lexikographie ist die KI ein Teilgebiet der Informatik, das sich mit der maschinellen Nachahmung menschlicher Intelligenz befasst, d.h. mit der Entwicklung intelligenter Systeme (Mainzer 2019), die bei der Durchführung unterschiedlicher Aufgaben Aspekte des menschlichen Verhaltens wie der Problemlösung oder der Entscheidungsfindung simulieren (McCarthy 2007). Die KI hat viele Anwendungsbereiche, dementsprechend spielt sie eine wesentliche Rolle bei vielen Prozessen und den dadurch geschaffenen Endprodukten, wie z.B. bei der Entwicklung autonomer Fahrzeuge oder Roboter (z.B. für die Erforschung des Weltraums), bei den medizinischen Diagnosen, bei der Marktforschung für ein Produkt, bei der Bilderkennung (Gesichtserkennung oder Objekterkennung), u.a. Die natürliche Sprachverarbeitung (NLP), die zur Datenverarbeitung und -interpretation der menschlichen Kommunikationsprozesse Computeralgorithmen und maschinelles Lernen heranzieht, gilt als Teilgebiet der KI. Als besonders relevant erweisen sich die Sprachmodelle, da sie in unterschiedlichen Bereichen eingesetzt werden können bzw. als Grundlage unterschiedlicher Produkte und Ressourcen dienen können. Im Weiteren werden einige genannt:

— Sprachassistenten wie *Alexa, Siri* oder *Google home*.
— Chatbots und virtueller Kundenservice: Es handelt sich um Softwareanwendungen bzw. virtuelle Assistenten, die häufig gestellte Fragen beantworten. Man strebt z.B. dabei an, den Kundenservice zu verbessern (Banken, Online-Shops, u.a.). Sie werden auch für den Querverkauf von Produkten, als Nachrichten-Bots beim elektronischen Handel oder als Nachrichten-Apps wie *Facebook Messenger*. Als Chatbots lassen sich ChatGPT (s. 3 und 4), *Bing CHAT* (https://www.bing.com/) oder *Bard* (https://bard.google.com/) nennen. Mit diesen Dienstleistungen und Werkzeugen stehen Untersuchungen zur Analyse der Meinungsforschung oder der Analyse von Gefühlen in engem Zusammenhang: ihre Ergebnisse finden eine direkte Anwendung in Gebieten wie dem elektronischen Handel oder der Meinungsanalyse in den sozialen Netzwerken, indem sich ein Text oder ein nutzergenerierter Inhalt durch eine automatische maschinelle Analyse als positiv, negativ oder neutral bewerten lässt. Einige Beispiele dazu sind *Linguakit* (https://linguakit.com/es/analizador-de-sentimient) oder *SentiWordNet* (https://github.com/aesuli/SentiWordNet).

— Roboter für die schriftliche Textproduktion, wie *z.B. Inferkit* (https://inferkit.com/), *Sassbook AI* (https://sassbook.com/), oder AI-Schreibsysteme, wie *Rytr* (https://app.rytr.me). Insgesamt erstellen sie automatisch Texte aus einer reduzierten Anzahl an Wörtern als Ausgangspunkt.
— Weitere Generatoren sind vorhanden:
  — Für die Erstellung von sportlichen Berichten, Zusammenfassungen, vereinfachten Texten usw. liegen Beispiele vor (Nallapati et al. 2016 oder Roemmele 2016).
  — Für die Wiedergabe mündlicher Sprache und Interaktionen im Gespräch, wie z.B. Systeme, die mit der Stimme einer Person trainiert werden und Text produzieren bzw. Gespräche führen. Es gibt Apps wie *Pi.ai/talk* (https://pi.ai/talk), *Call Annie* (https://callannie.ai) oder *character.ai* (https://beta.character.ai/; für Jugendliche), eine Art KI-Freunde bzw. intelligente Assistenten.
  — Für die Erstellung von Bildern nach Beschreibungen in natürlichen Sprachen, wie *DALL-E* (https://openai.com/dall-e-2), von Open AI.
  — Automatische Übersetzung, wie z.B. *WIPO Pearl* (https://wipopearl.wipo.int/en/linguistic).

Unausgesprochen weisen die KI und die Lexikographie andere Endprodukte und Ziele auf. Eine weitere Entscheidung sei hier hervorzuheben: Sprachmodelle streben an, dass Maschinen nachahmen, wie Menschen Wörter verwenden, d.h. die menschliche Kommunikation; die KI hingegen hat die Entwicklung von Maschinen und KI-Tools als Ziel, die intelligent agieren. Die Wechselwirkungen sind nicht zu übersehen.

### 3.   Effizienz und Intelligenz am Beispiel von ChatGPT

An erster Stelle scheint es sinnvoll, zu erläutern, warum sich der vorliegende Beitrag mit ChatGPT befasst und warum wir ChatGPT als intelligentes System auswählen. Benutzungsstudien (vgl. Domínguez Vázquez und Valcárcel Riveiro 2015 oder Müller-Spitzer und Koplenig 2015) stellen fest, dass BenutzerInnen Faktoren wie (a) eine schnelle und bequeme Abfrage sowie (b) einen leichten und kostenfreien Zugang schätzen. ChatGPT erfüllt diese Voraussetzungen, und deswegen kann vorausgesagt werden, dass BenutzerInnen es einem Wörterbuch bevorzugen würden. Dies lässt sich auch dadurch begründen, dass ChatGPT in nur fünf Tagen die Grenze von einer Million NutzerInnen überschritten hat (s. Abbildung 1).

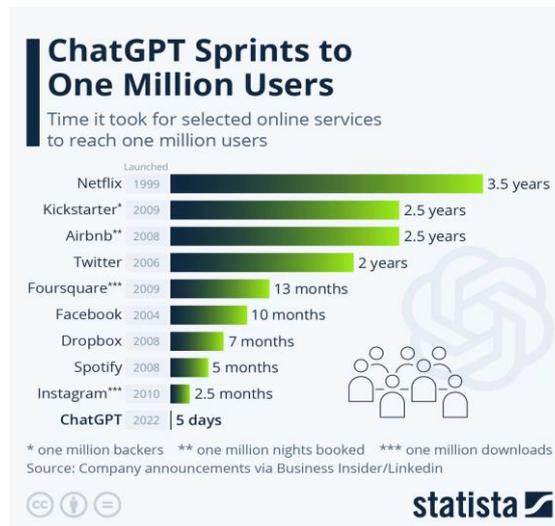

**Abbildung 1:**   Zeit, die Online-Dienste bis zum Erreichen von 1 Million Nutzer-Innen benötigten

ChatGPT ist ein Chatbot, der aus einem Sprachmodell mit generativen Vortraining stammt und ein autoregressives System enthält, d.h., dass es Tokens vorhersagt, sie zum Prompt hinzufügt und sie wieder in das Modell einspeist. Das ist der Grund, weshalb sich solche Modelle als sehr effizient erwiesen haben, wenn die Eingabeaufforderung darin besteht, Texte oder Textsorten zu (re)produzieren (vgl. Radford et al. 2018). Die Grundannahme der hier zu präsentierenden Pilotexperimente basiert darauf, dass ChatGPT — wie andere generative Sprachmodelle — menschenähnliche Texte von hoher Qualität generieren kann. Ob das in der Tat bei der Erstellung von Wörterbuchartikeln im Deutschen und im Galicischen so ist, ist Gegenstand dieser Untersuchung (s. 4).

In diesem Zusammenhang sei hier auf den Intelligenzbegriff hinzuweisen, der laut der Wikipedia-Definition mit der Anwendung von kognitiven Fähigkeiten zusammenhängt:

> Intelligenz (von lateinisch *intellegere* „erkennen", „einsehen"; „verstehen"; wörtlich „wählen zwischen …" von lateinisch *inter* „zwischen" und *legere* „lesen, wählen") ist die kognitive bzw. geistige Leistungsfähigkeit speziell im Problemlösen. Der Begriff umfasst die Gesamtheit unterschiedlich ausgeprägter kognitiver Fähigkeiten zur Lösung eines logischen, sprachlichen, mathematischen oder sinnorientierten Problems.

Hier kommt die Frage auf, was das Wort „Intelligenz" im Konzept „künstliche Intelligenz" zu bedeuten hat. Dazu äußert sich McCarthy (2007: 2) wie folgt:

> It is the science and engineering of making intelligent machines, especially intelligent computer programs. It is related to the similar task of using computers to understand human intelligence, but AI does not have to confine itself to methods that are biologically observable.

McCarthy (2007: 2) beschreibt diese Intelligenz als die Fähigkeit, bestimmte Ziele zu erreichen. Diese Beschreibung bringt eine weitere Frage mit sich, und zwar, was intelligente Maschinen sind oder inwiefern sie als solche aufgefasst werden können. Darauf antwortet McCarthy (2007: 3), dass die Maschinen nicht intelligent sind, aber er ergänzt seinen Ansatz wie folgt:

> Intelligence involves mechanisms, and AI research has discovered how to make computers carry out some of them and not others. If doing a task requires only mechanisms that are well understood today, computer programs can give very impressive performances on these tasks. Such programs should be considered "somewhat intelligent".

Da der Intelligenzbegriff mit dem Verfahren zur Problemlösung in engem Zusammenhang steht, ist aus computergestützter Perspektive auf die Verbindung zwischen Intelligenz und System näher einzugehen. Folglich definiert Mainzer (2019: 3) intelligente Systeme folgenderweise:

> Ein System heißt intelligent, wenn es selbständig und effizient Probleme lösen kann. Der Grad der Intelligenz hängt vom Grad der Selbstständigkeit, dem Grad der Komplexität des Problems und dem Grad der Effizienz des Problemlösungsverfahrens ab.

Auf die drei genannten Intelligenzgrade eines Systems wird jetzt eingegangen:

(a) Der Grad der Selbständigkeit hängt von dem zu benutzenden System ab. Eine Computeranwendung kann selbständig sein, wenn sie unabhängig von anderen Ressourcen Probleme lösen kann.
(b) Der Grad der Komplexität des Problems hängt von der Handlung selbst ab. In diesem Sinne ist mit einer bunten Vielfalt von Situationen zu rechnen, die den Komplexitätsgrad erhöhen können. Hier geht man davon aus, dass die Benutzenden ein System zu Rate ziehen sollen, das zur Beantwortung ihrer Anfragen geeignet ist; z.B. ist von einem Korpus nicht zu erwarten, dass es Definitionen liefert, wie von einem Wörterbuch nicht verlangt werden kann, dass es automatisch Texte generiert.
(c) Der Grad der Effizienz des Lösungsverfahrens hängt davon ab, ob ein System in der Tat ein konkretes Problem lösen bzw. eine konkrete Frage angemessen beantworten kann. Bei der Bestimmung der Effizienz sind der Wirkungsgrad (s. 5) und die Rolle der BenutzerInnen (s. 5) von entscheidender Bedeutung.

Seitens der Lexikographie wird der Effizienzbegriff klar definiert. Laut Wiegand (1998: 259) ist die Benutzungseffizienz als Resultat einer Adäquatheit des Produk-tes (des Wörterbuches) an die Bedürfnisse der BenutzerInnen zu verstehen, wozu die wissenschaftlichen Kenntnisse der Benutzenden und die Etablierung der Konsultation eines Referenzwerkes als kulturelle Praxis einen großen Beitrag leisten. Im Grunde genommen haben beide — die Lexikographie und die KI — ein gemeinsames Prinzip: Man ist auf der Suche nach Lösungen für ein Problem und benötigt dazu Mechanismen und Strategien. Im Gegensatz zu Menschen stützen sich Maschinen jedoch auf Theorien menschlicher Denk- und Lernprozesse, und insbesondere auf die Nachbildung dieser Prozesse (vgl. 5). In diesem Zusammenhang sei daran zu erinnern, dass die KI eine unvorstellbare Menge an Daten analysieren kann und in der Lage ist, Muster und Daten auf der Grundlage menschlicher Kommandos zu reproduzieren. Die konkrete Frage lautet dann, inwiefern ein Gespräch mit einem intelligenten System effizient zur Lösung eines sprachlichen bzw. lexikographischen Problems führen kann, und das hängt vom Effizienzgrad des Lösungsverfahrens ab. Den Effizienzbegriff verstehen wir daher in der vorliegenden Arbeit auf zwei Weisen: zum einen spricht die Datenmenge, über die ein System oder ein Nachschlagewerk verfügen, gewissermaßen für einen höheren Wirkungsgrad (s. 5.4); zum anderen gilt der Benutzertyp als Voraussetzung für die Effizienz des Problemlösungsverfahrens (s. 5.1 und 5.4). Begriffe wie „Handlung" oder „Handelnde" haben unseres Erachtens Einfluss auf die Bestimmung des Intelligenzbegriffes und auf die Problemlösung (s. 5).

## 4.     Pilotexperimente mit ChatGPT

### 4.1     Zum Stand der Forschung

Seit der Entstehung von ChatGPT im November 2022 sind mehrere Untersuchungen veröffentlicht worden, die sich mit der Umsetzung dieses Chatbots im Bereich der Lexikographie befassen. Einige dieser Studien (vgl. de Schryver und Joffe 2023; Rundell 2023; Jakubíček und Rundell 2023) vertreten die Ansicht, dass eine menschliche Bewertung und Überprüfung bei der Aufführung von lexikographischen Tätigkeiten immer noch erforderlich seien, obschon Maschinen die harte Arbeit — nämlich die Programmierung und Formatierung lexikographischer Daten — bereits übernehmen können. Andere Untersuchungen (vgl. Nichols 2023; Barrett 2023, zit. nach de Schryver 2023) deuten darauf hin, dass ChatGPT Informationen bieten könnte, die nicht vollkommen zuverlässig sind und Verbesserungen erfordern, und dass das System zum Teil „halluziniere".

Aus quantitativer Sicht (vgl. Phoodai und Rikk 2023) vermag ChatGPT jedoch eine bessere Leistung vorzuzeigen, denn das System ist in erster Linie dazu trainiert, Textlücken zu erfüllen.

Die erwähnten Arbeiten antworten teilweise auf die Fragen, die in der Einführung (s. 1) gestellt wurden. Außerdem kann zum einen festgestellt werden, dass lexikographische Kenntnisse immer noch von Vorteil sind, weil ExpertInnen im Bereich der Lexikographie diejenigen sind, die die Aufgabe übernehmen, die Antworten von ChatGPT auf lexikographisch-orientierte Fragen auszuwerten (vgl. de Schryver 2023; Alonso-Ramos 2023). In dieser Hinsicht ist anzumerken, dass der Mangel an Untersuchungen über die Handlungen der BenutzerInnen *in-actu* eklatant ist[4]. Zum anderen lässt sich feststellen, dass ChatGPT in der Lage ist, lexikographische Tätigkeiten auszuführen (vgl. de Schryver und Joffe 2023; Tran et al. 2023), wenn auch die Präzision und die Qualität der Ergebnisse etwas zu wünschen übrig lassen.

Daraus lässt sich schließen, dass die Benutzenden zu bewerten haben, wie kohärent der Output der Maschine ist. Laien hätten noch Schwierigkeiten, ein Prompt genauer zu erstellen. Im Falle einer lexikographischen Anfrage müssen sie noch dazu über die notwendigen spezifischen Kenntnisse verfügen. Hingegen können ExpertInnen nicht nur zur Evaluation und zur akribischeren Prompterstellung beitragen, sondern auch zur Verbesserung des Outputs. Das knüpft an die Frage nach der Effizienz und Qualität bei der Generierung lexikographischer Einträge seitens ChatGPT an.

Als eigene Textsorte müssen Wörterbuchartikel[5] — wie jeder Text — Kohärenz und Kohäsion aufweisen. Das bedeutet, dass der Output von ChatGPT einen gewissen Grad an inhaltlicher Verbindung und an textueller Organisation aufweisen muss. Wiegand (1996) erklärt, dass Wörterbuchartikel Teil des lexikographischen Textes sind und sich wiederum in Textteile (wie z.B. Lemmazeichengestaltangabe oder Bedeutungsangabe) segmentieren lassen. Davon ausgehend stellen wir die Frage, inwiefern ChatGPT als zur Erstellung von Texten trainiertes Sprachmodell einen lexikographischen Text mit einem bestimmten Grad an Kohärenz und Qualität (also Effizienz) verfassen kann. Im Folgenden wird das statistische Verfahren (s. 4.2) vorgestellt und die Outputs der Pilotexperimente (s. 4.3 und 4.4) angesichts ihres lexikographischen Wertes diskutiert.

### 4.2 Zur Methode und zur statistischen Auswertung

Zur Beantwortung der Frage, ob und inwiefern ChatGPT kohärente lexikographische Texte produziert, sind zwei ChatGPT-Sessions für jede der beiden Sprachen durchgeführt worden. Um die Ergebnisse quantitativ und qualitativ vergleichen zu können, haben wir nur ein Prompt ausgewählt, denn wir streben nicht an, das System zu trainieren oder den Prozess bis zur Erstellung eines zufriedenstellenden endgültigen Wörterbuchartikels zu evaluieren. Das vorgeschlagene Verfahren hat noch keine Aufmerksamkeit in der Fachliteratur gefunden, es entspricht aber der Mehrzahl der möglichen Verwendungen dieses Chatbots (s. Abbildung 1).

Die Auswahl des Deutschen und des Galicischen als Arbeitssprachen zielt darauf ab, andere Sprachen als das Englische zu prüfen, denn die meisten dem System zugrundeliegenden Daten, mehr als 90% (vgl. Rundell 2023), sind lediglich auf Englisch verfügbar. Es wird dadurch angestrebt, Einsichten in die Datenbasis zu gewinnen, auf der das Sprachmodell ChatGPT trainiert wurde.

Analysiert wird konkret die ChatGPT-Antwort auf den Prompt „Erstelle einen Wörterbuchartikel für L". Im galicischen Experiment musste zu dem Prompt hinzugefügt werden, dass das Ergebnis auf Galicisch erfasst werden sollte („Crea un artigo lexicográfico para L en galego"), weil sonst der Output oft auf Portugiesisch oder eben auf Spanisch erfolgt.

Die Abfragen auf Deutsch wurden für folgende Lemma formuliert: *abgeben*, *abholen*, *abschließen*, *anbieten* und *anmachen*. Es handelt sich um die fünf ersten trennbaren transitiven Verben der Goethe-Wortschatzliste zu A2. Für die Analyse der Ergebnisse auf Galicisch wurden die fünf im CORGA-Korpus häufigsten transitiven Verben ausgewählt: *facer*, *dicir*, *dar*, *saber* und *querer*.

Die in dieser Studie angewandten statistischen Methoden basieren auf der Verarbeitung von *n*-Grammen[6]. Zum quantitativen Vergleich der Outputs der zwei ChatGPT-Sessions werden zwei Metriken[7] umgesetzt: *Jaccard* und *Dice*[8]. Zum Vergleich der Outputs von der ersten Session in ChatGPT mit dem jeweiligen Wörterbuchartikel aus Referenzwörterbüchern wird eine dritte Metrik –*ROUGE-N*– herangezogen. Als Referenzwörterbücher gelten hier für das deutschsprachige Experiment der DUDEN und für das Experiment mit dem Galicischen das DRAG. Im Folgenden werden die Metriken sowie die erzielten Ergebnisse angeführt:

— Der *Jaccard*-Ähnlichkeitskoeffizient dient als Metrik zur Bewertung der Ähnlichkeit zwischen zwei Sets von *n*-Grammen und wird als Verhältnis der Anzahl gemeinsamer *n*-Gramme zur Gesamtzahl der Elemente beider Sets definiert. Der Text, der als Input gegeben wird, muss durch Leerzeichen tokenisiert werden und die Tokens werden in Sets umgewandelt, welche die statistische Berechnung ermöglichen. Der *Jaccard*-Koeffizient ergibt sich aus der Bestimmung von Schnittmengen und der Vereinigung der Sets. Der *Jaccard*-Koeffizient wird hier als Prozentsatz ausgedrückt.
— Der *Dice*-Koeffizient dient ebenso als Metrik zur Messung der Ähnlichkeit zwischen zwei Mengen. Zunächst werden die Texte durch Leerzeichen tokenisiert und in Vektoren von Wörtern umgewandelt. Diese Wörter werden dann in Sets transformiert, wodurch doppelte Wörter entfernt werden. Der *Dice*-Koeffizient wird berechnet, indem man die Anzahl der gemeinsamen Elemente verdoppelt und durch die Gesamtanzahl der Elemente beider Sets teilt. Der resultierende Koeffizient liegt zwischen 0 und 1, wobei 0 keine Ähnlichkeit und 1 vollständige Ähnlichkeit bedeutet. Das Ergebnis wird schließlich als Prozentsatz ausgedrückt.

Tabelle 1 gibt einen umfassenden Einblick in die statistischen Ergebnisse beider Verfahren. Die Prozentwerte verdeutlichen, in welchem Ausmaß ein Ähnlichkeitsverhalten zwischen den Outputs der beiden ChatGPT-Sitzungen festzustellen ist. Jede Zeile in der Tabelle bezieht sich auf die Wörterbuchartikel, die ChatGPT in den beiden Sitzungen für jedes ausgewählte Lemma generiert hat.

Als Beispiel für eine angemessene Interpretation der Daten lässt sich beobachten, dass im Fall des Lemmas *abgeben* eine Ähnlichkeit von 15,67% (*Jacard*) und 27,09% (*Dice*) zwischen dem Wörterbuchartikel der Sitzung 1 und dem der Sitzung 2 besteht. Außerdem wird in der letzten Zeile der Durchschnitt der Ähnlichkeit für jede Metrik und für jede Sprache angegeben.

|  | **Pilotexperiment: Deutsch** | |  | **Pilotexperiment: Galicisch** | |  |
| --- | --- | --- | --- | --- | --- | --- |
|  | *Jaccard* | *Dice* |  | *Jaccard* | *Dice* |  |
| *abgeben* | 15,67% | 27,09% |  | 14,21% | 24,88% | *facer* |
| *abholen* | 22,40% | 36,61% |  | 14,01% | 24,59% | *dicir* |
| *abschließen* | 20,79% | 34,42% |  | 12,89% | 22,83% | *dar* |
| *anbieten* | 20,39% | 33,86% |  | 12,50% | 22,22% | *saber* |
| *anmachen* | 19,30% | 32,35% |  | 13,19% | 23,3% | *querer* |
|  | **19,71%** | **32,87%** | **Durchschnitt** | **13,36%** | **23,56%** |  |

**Tabelle 1:** Ähnlichkeitsverhalten zwischen den Outputs der zwei ChatGPT-Sitzungen

Die Ergebnisse weisen eine konsistente Tendenz auf, bei der der *Jaccard*-Koeffizient stets niedrigere Werte im Vergleich zum *Dice*-Koeffizienten anzeigt. Daraus ergibt sich, dass die Distanz zwischen den *Jaccard*- und *Dice*-Werten konstant oder zumindest ähnlich ist. Die konsistente Beziehung zwischen den beiden Koeffizienten bedeutet, dass die statistischen Verfahren angemessen angewandt worden sind. Auf der Grundlage des Pilotexperiments mit der deutschen Sprache stellt sich heraus, dass die erzielten Prozentsätze höher ausfallen, und dies veranschaulichen bereits die durchschnittlichen Ergebnisse: Beim Pilotexperiment mit dem Deutschen liegen die Werte bei 19,71% (*Jaccard*) und 32,87% (*Dice*), während die Werte im Experiment mit dem Galicischen 13,36% (*Jaccard*) und 23,56% (*Dice*) betragen. Die höheren Ergebnisse könnten mit einer umfassenderen Datengrundlage zusammenhängen, die für eine präzisere Modellierung und Textproduktion sorgt.

Aus dieser quantitativen Analyse lässt sich ebenfalls folgern, dass der Effizienzgrad von ChatGPT gering ist. Dies ist auf mehrere Faktoren zurückzuführen. Ein entscheidender Faktor ist die mangelnde Konstanz im generierten

Output. Die Unbeständigkeit in den erzeugten Texten kann zu Inkonsistenzen in der Qualität führen und erschwert eine zuverlässige Leistungsbewertung (s. 5.3). Dies könnte auf Schwächen im Trainingsprozess oder in den Modellparametern hinweisen, die zur Gewährleistung einer stabilen und konsistenten Ausgabe eine Überprüfung und Anpassung erfordern. Dieses Verfahren ermöglicht dennoch Einblicke in die Trainingsdaten, wobei man feststellen kann, dass die Daten, mit denen ChatGPT trainiert wurde, über weniger lexikographischen Text bzw. lexikographischen Textsorten verfügen. Außerdem müssen Benutzende den vom ChatGPT generierten Output bewerten (können), was zu einem erhöhten zeitlichen Aufwand führt und was die Benutzerfreundlichkeit des Systems beeinträchtigen kann. Die quantitativen Ergebnisse deuten zusammenfassend bei den beiden analysierten Sprachen darauf hin, dass ChatGPT bei lexikographischen Anfragen nicht präzise genug ist.

Die qualitativen Ergebnisse der *Jaccard*- und *Dice*-Metriken sind nicht das einzige entscheidende Bewertungskriterium, denn neben diesem wird die *ROUGE-N*-Methode eingesetzt, die eine effektive quantitative Auswertung der automatischen Textgenerierung im Vergleich zu Referenztexten ermöglicht[9]. Die *ROUGE-N*-Methode bietet eine Evaluierung der Textübereinstimmung an, insbesondere im Kontext der automatischen Textzusammenfassung. Hierbei erfolgt eine Tokenisierung der Texte in *n*-Gramme, die als Sets repräsentiert werden, um Duplikate zu eliminieren. Nach der Berechnung mathematischer Algorithmen fungieren die resultierenden F1-Score-Werte als quantitatives Maß für die Qualität automatisch von ChatGPT generierter Texte im Vergleich zu lexikographischen Referenztexten (in diesem Fall dem DUDEN und dem DRAG entnommen). Die Berechnung erfolgt durch die Überlappung von Unigrammen zwischen den Texten, wobei die *ROUGE*-1-Metrik F1-Score für die Ähnlichkeit ermittelt wird. Dies dient als Kriterium zur Beurteilung der Qualität automatisch generierter Texte. Zur Berechnung der *n*-Gramme und der Metriken werden die gesamten Wörterbuchartikel des DUDEN und DRAG genommen[10].

Tabelle 2 bietet einen Gesamtüberblick über die statistischen Ergebnisse der Textübereinstimmung zwischen dem von ChatGPT generierten lexikographischen Text und den Wörterbuchartikeln aus den ausgewählten Referenzwörterbüchern.

Die Ergebnisse der *ROUGE-N*-Metrik zeigen, dass die Ähnlichkeit zwischen den generierten ChatGPT-Texten und den Referenztexten gering ist, wobei nahezu so gut wie keine Übereinstimmung festzustellen ist. Der durchschnittliche Wert bei *ROUGE-N* für das Deutsche ist eine Übereinstimmung zwischen dem Wörterbuchartikel aus ChatGPT und dem aus Duden von 5,6%; im Falle des Galicischen beträgt sie 6,46%. Diese niedrigen *ROUGE-N*-Werte könnten mit Schwächen im Modelltraining, in den Daten oder in den Parametern zusammenhängen, die die Fähigkeit des Systems beeinträchtigen, relevante lexikographische Informationen adäquat zu extrahieren und wiederzugeben.

|  | **Pilotexperiment: Deutsch** |  | **Pilotexperiment: Galicisch** |  |
|---|---|---|---|---|
|  | ROUGE-N |  | ROUGE-N |  |
| *abgeben* | 4,95% |  | 5,27% | *facer* |
| *abholen* | 6,27% |  | 7,02% | *dicir* |
| *abschließen* | 4,72% |  | 4,76% | *dar* |
| *anbieten* | 6,90% |  | 6,67% | *saber* |
| *anmachen* | 5,17% |  | 8,58% | *querer* |
|  | **5,60%** | **Durchschnitt** | **6,46%** |  |

**Tabelle 2:** Übereinstimmung zwischen dem generierten lexikographischen ChatGPT-Text und den Wörterbuchartikeln aus Referenzwörterbüchern

Insgesamt lässt sich festhalten, dass die *Jaccard*-, *Dice*- und *ROUGE-N*-Werten niedrig sind und darüber hinaus, dass das System aus quantitativer Sicht nicht effizient ist (s. 5.4). Dies scheint auf eine potenzielle Begrenzung des lexikographischen Wissens des Systems hinzuweisen, da die niedrigen Prozentwerte auf eine geringe Übereinstimmung zwischen den generierten Texten und den erneut generierten Texten oder den Referenztexten hindeuten. Ein weiterer Aspekt, der in Betracht gezogen werden sollte, ist die Berücksichtigung der gesamten Mikrostruktur der Texte anstelle spezifischer Textabschnitte. Diese Herangehensweise ermöglicht eine umfassendere Bewertung der generierten Inhalte und kann zu niedrigeren prozentualen Ergebnissen im Vergleich zu dem Fall führen, in dem ein spezifischer Textteil (wie z.B. die Bedeutungsangabe oder die Beispielangabe) isoliert ausgewählt worden ist.

Weiterhin kann neben den quantitativen Ergebnissen eine qualitativ ausgerichtete Analyse der Textteile einen tieferen Einblick in den Effizienzgrad des Systems gewährleisten. Dazu setzen wir uns in 4.3 und 4.4 mit der Bewertung von spezifischen Textteilen und Aspekten der generierten Texte wie Kohärenz oder Relevanz auseinander.

### 4.3 ChatGPT als Verfasser von Wörterbuchartikeln im Deutschen

Zur Auswertung des Inhalts und der Outputs[11] achten wir auf die unterschiedlichen Textteilen — lexikographischen Angaben — in der Antwort von ChatGPT. Konkret setzen wir uns mit der Analyse der Angaben auseinander, die nach

Engelberg und Lemnitzer (2009: 135) wie folgt definiert werden: „die funktionalen Textsegmente, die in einer Angabebeziehung zu bestimmten Elementen [...] stehen und die es dem Benutzer ermöglichen sollen, aus ihnen bestimmte Informationen über den Wörterbuchgegenstand zu gewinnen."

Die hier diskutierten Angabeklassen beziehen sich grundsätzlich auf die inhaltliche Form der ausgewählten sprachlichen Einheiten. Die Textsegmente, deren Funktion darin besteht, die innere Zugriffsstruktur des lexikographischen Werkes deutlicher zu konzipieren, werden ausgeklammert. Die Ergebnisse lauten wie folgt:

(i) Lemmazeichengestaltangabe: Diese Angabe kommt in den zwei Sessions in derselben Form vor („Wörterbuchartikel: *L*"). Anzumerken ist dennoch, dass in der ersten ChatGPT-Session diese Angabe größer gedruckt wird.

(ii) Wortartangabe: Diese Angabe tritt in den zwei Sessions auf und ist immer richtig vorgegeben, denn die ausgewählten Wörter gehören zur Wortart der Verben. In der ersten Session wird diese Angabe fettgedruckt und in der zweiten Session ist die Wortartangabe kursiv hervorgehoben.

(iii) Ausspracheangabe: Diese Angabe wird nur in der zweiten Session präsentiert und erfolgt durch die Verwendung der IPA-Transkription, die zudem mit dem entsprechenden Akzent versehen wird. Die Darstellung der phonetischen Transkription ist richtig. Die Tatsache, dass diese Angabe nicht in jeder Session vorgestellt wird, lässt sich als eine der Schwächen des Systems verstehen.

(iv) Bedeutungsangabe: Die Bedeutungen und die verschiedenen Lesarten der ausgewählten Lemmata sind in den zwei Sessions vorhanden, aber werden in unterschiedlicher Form präsentiert. In der ersten Session werden zunächst sehr allgemeine Informationen angeboten: „Das Verb L hat verschiedene Bedeutungen, abhängig vom Kontext, in dem es verwendet wird". Erst nach dieser wenig aussagekräftigen Paraphrase werden die unterschiedlichen Lesarten aufgelistet, aber die Formulierungen bedürfen einer menschlichen Überprüfung, da diese gelegentlich wenig semantischbasiert sind. Folgendes Beispiel veranschaulicht dies: die dritte Lesart des Verbs *abholen* lautet „jemandem von einem Ort abholen".

In der zweiten Session wird keine allgemeine Definition vermittelt und die Lesarten werden durchnummeriert und vorgestellt. Die Formulierungen der Definitionen sind auch Kritik bedürftig, weil die Semantik hier wieder im Hintergrund steht. Ein Beispiel dafür ist die fünfte Lesart des Verbs *abschließen*: „etwas beenden, indem man es abschließt [...]". Im Fall der zweiten Session kommt als letzter Textteil ein Abschnitt vor, der „Verwendung" genannt wird und darunter fallen allgemeine Erklärungen hinsichtlich der Bedeutung der Lemmata: „Das Verb L wird in verschiedenen Kontexten verwendet [...]. Es findet Anwendung im Alltag, in formellen und informellen Gesprächen, im Handel, im Geschäftsleben, im Tourismus und vielen anderen Bereichen des täglichen Lebens". Solche allgemeinen Erläuterungen sind auch in der ersten Session unter

       der Überschrift „zusätzliche Informationen" vorhanden: „Das Verb L ist in der deutschen Sprache allgemein gebräuchlich und vielseitig einsetzbar".
(v) Flexionsangabe: Diese Angabe wird ausschließlich während der ersten Session erstellt. Dabei liegt der Fokus auf der Darstellung des Flexionsparadigmas der Verben im Präsens und Präteritum. In Wörterbüchern wird normalerweise das Flexionsparadigma nicht in seiner vollständigen Form angegeben, sondern lediglich durch drei Formen repräsentiert: Präsens (3. Person Singular), Präteritum (3. Person Singular), Partizip Perfekt.
(vi) Beispielangabe: Beispielsätze, die die Verwendung der Lemmata veranschaulichen, werden in den zwei Sessions vorgelegt, aber ihre Darstellung weicht in den beiden Sessions voneinander ab. In der ersten Session werden Beispielsätze unter der Überschrift „Verwendungsbeispiele" angeboten, aber sie werden nicht einer bestimmten Lesart zugeordnet. Bei jeder Lesart ist auch jeweils ein Beispielsatz vorhanden. In der zweiten Session hingegen tritt unter jeder Lesart ein Beispielsatz auf, der das kombinatorische Verhalten des Lemmas darstellt. Die Generierung dieser Beispielsätze ist oft wenig hilfreich und erklärend, da die Kombinatorik der Lemmata nicht deutlich dargestellt wird. Es kommen Beispielsätze wie folgende vor: „Sie hat sich in ihn verliebt und macht ihm Avancen" (Session 1, Le20mma: *anmachen*)[12] oder „Der Klempner muss die Rohre richtig abschließen, um Lecks zu vermeiden". (Session 2, Lemma: *abschließen*).
(vii) Synonymen- und Antonymenangaben: In beiden Sessions werden Synonyme und Antonyme zu den entsprechenden Lemmata angegeben, aber es erscheinen meist unterschiedliche Synonyme und Antonyme bei jeder Session, und deren Anzahl ist in der zweiten Session höher als in der ersten. Sogar Funktionsverbgefüge erscheinen als Synonyme.
(viii) Wortfamilienangabe: In der zweiten Session kommt ein Textteil vor, der „verwandte Begriffe" genannt wird. Dort werden immer zwei Substantive angegeben, die von dem entsprechenden Verb abgeleitet sind. Das erste Nomen verweist immer auf die durch das Verb beschriebene Handlung („Abgabe", „Abholung", „Abschluss" usw.), während sich das zweite Substantiv auf eine Person bezieht, die die Handlung vollzieht („Abgebender", „Abholer", „Anbieter" usw.).
(ix) Etymologische Angabe: Diese Informationen werden in den zwei Sessions geliefert, aber die vermittelten Informationen sind unterschiedlich. Als Beispiel dient der Fall des Verbs *abgeben*, das der ersten Session zufolge aus dem Präfix *ab* und dem althochdeutschen *gēben* stammt und das laut der zweiten Session dem mittelhochdeutschen *abegeben* entstammt. In der ersten Session wird der Ursprung der Verben auf das Althochdeutsche zurückgeführt, wogegen der Gebrauch derselben Lemmata laut der zweiten Session erst ab dem 16. und 17. Jahrhundert dokumentiert wurde.

Insgesamt lässt sich schließen, dass die angebotenen Informationen nicht konstant sind und dass sie sich teilweise widersprechen. Das deutet auf einen geringen Effizienzgrad des Systems hin, denn zur Erschließung sprachlicher Kenntnisse dank des generierten lexikographischen Texts ist eine Überprüfung in anderen Quellen erforderlich. Ein Vergleich mit den entsprechenden Wörterbuchartikeln im DUDEN weist ebenso darauf hin, dass die Mikrostruktur der lexikographischen Artikel im Referenzwörterbuch fixiert ist und dass dieselben Textteile in derselben Reihenfolge auftreten. ChatGPT bietet Informationen an, die zum Teil von denen im DUDEN abweichen.

**4.4   ChatGPT als Verfasser von Wörterbuchartikeln im Galicischen**

Kommen wir nun zum Experiment mit Galicisch als Metasprache[13]. Um es vorwegzunehmen: Das System von ChatGPT betrachtet das Galicische nicht immer als eigenständige Sprache, denn Textteile in der Antwort werden meist auf Portugiesisch und zum Teil auch auf Spanisch verfasst.

Auffällig beim Pilotexperiment mit Galicisch ist, dass im Output immer eine Einführung und eine Schlussfolgerung stehen. Offenbar hat ChatGPT das Wort „Artikel" hier im Sinne von ‚Aufsatz' verstanden, und deswegen neigt das System dazu, die Informationen ohne Kondensierung anzuordnen und mit einer gewissen Struktur vorzustellen. In den zwei Sessions kommen zum Beispiel Einführungen wie die folgende zustande:

> A palabra *L* é unha das palabras máis poderosas e frecuentes na lingua galega. Co seu uso tan común, é esencial comprender a súa ampla gama de significados e usos. Neste artigo lexicográfico, exploraremos a palabra L en galego, analizando as súas múltiples acepcións, construcións gramaticais e exemplos de uso.[14] (Session 1)

Nach der Beschreibung der Textteile, auf die wir gleich eingehen werden, kommen Schlussfolgerungen wie die folgende vor:

> O verbo L é um elemento central na língua galega devido à sua versatilidade e importância na comunicação. Sua conjugação e usos variados tornam-no uma ferramenta essencial para expressar uma ampla gama de ações, causas e estados de coisas. É uma palavra que transcende a gramática e se torna uma parte integrante da vida cotidiana para os falantes galegos, contribuindo para a riqueza e diversidade desta língua.[15] (Session 2)

Folgende mikrostrukturelle Textteile treten in den generierten galicischen Wörterbuchartikeln von ChatGPT vor:

(i)     Lemmazeichengestaltangabe: Diese Angabe kommt in den zwei Sessions in Form einer kreativen Überschrift vor. In der Session 1 werden die Wörterbuchartikel mit einem Titel versehen (übersetzt: „L – eine lexikographische Analyse auf Galicisch"), während die Überschriften in der zweiten Session eher kreativ und attraktiv sind (übersetzt: „*L* – das vielseitige Verb der galicischen Sprache").

(ii) Bedeutungsangabe: In der ersten Session werden die Lesarten der Lemmata durchnummeriert und getrennt angegeben. Die erste Lesart wird „primäre Bedeutung" genannt. Die Definitionen beginnen immer mit der Formulierung „Bedeutung von…", was in der lexikographischen Praxis nicht üblich ist. In der zweiten Session wird eine einzige Definition angegeben, die normalerweise auf mehrere Lesarten Bezug nimmt. Bei der Erfassung der Definitionen werden häufig wenig aussagekräftige Informationen geliefert: „É um dos verbos mais comuns e essenciais na língua galega, desempenhando um papel fundamental na comunicação e transmissão de informações"[16]. Hinzu sollte noch ergänzt werden, dass die Lesarten der Lemmata in der Session 2 unter der Überschrift „häufige Verwendungen" verzeichnet werden.

(iii) Flexionsangabe: Diese Angabe kommt in beiden Sessions vor; in der ersten Session wird erklärt, dass die Verben in allen Tempora konjugiert werden können. Beispiele von vier Tempora (Präsens, zwei Formen des Präteritums, Futur) werden geboten. In der zweiten Session hingegen werden die Verben im Präsens vollständig konjugiert.

(iv) Idiomangabe: In der ersten Session ist ein Textteil vorhanden, in dem idiomatische Redewendungen aufgelistet werden. Eine menschliche Überprüfung zeigt aber, dass nicht alle als Idiome aufgefasst werden können. Beispiele davon sind „saber **a algo**" (‚**nach etwas** schmecken') oder „dicir unha mentira piadosa" (‚eine Notlüge erzählen').

(v) Beispielangabe: Es wird kein Textteil erzeugt, in dem Beispielsätze systematisch den Gebrauch eines Lemmas beschreiben. Auf jede Lesart folgt aber in der ersten Session eine Liste von „Beispielen", die teilweise die Verwendung der lexikalischen Einheiten veranschaulichen. Einige dieser Beispiele fungieren unserer Ansicht nach als Kollokationen in der galicischen Sprache: „dar un paseo" (‚einen Spaziergang machen') oder „dar consello" (‚Ratschläge geben').

(vi) Angabe der diatopischen Variation: In der zweiten Session werden Informationen hinsichtlich der diatopischen Variation gegeben, aber es handelt sich um keine genauen Informationen, die Bezug auf die Variation eines bestimmten Lemmas nehmen. Dieser Teil hält fest, wie varietätenreich die galicische Sprache ist.

Vergleicht man diese Informationen mit der Struktur des DRAG, stellt man fest, dass beim Output keine Systematisierung hinsichtlich der Anordnung, Gliederung und Beschreibung der Textteile existiert. Ähnlich wie im Fall des deutschen Experiments fehlt es an konstanten Informationen, was zu einem mangelnden Effizienzgrad führt. Die Unbeständigkeit oder das Fehlen von konstanten bzw. kohärenten[17] Daten beeinträchtigt maßgeblich die Wirksamkeit des Systems, da eine zuverlässige Grundlage für Entscheidungen in der lexiko-graphischen Beschreibung fehlt. Dieser Mangel an Stabilität kann potenziell zu Unsicherheiten und ineffizienten Abläufen auf der Seite der BenutzerInnen führen.

**5.        Gesamtanalyse der Ergebnisse**

Im letzten Abschnitt sollen die Hauptergebnissen der Untersuchung ausgeführt werden, die unausgesprochen auf die Unterscheidung zwischen der Lexikographie und den Sprachmodellen bzw. der künstlichen Intelligenz hindeuten.

**5.1     Handlung und Handelnde**

Einige Inkonsistenzen bei den Ergebnissen von ChatGPT lassen sich durch die Handlung und die Handelnden erläutern. Beim Erstellungsprozess eines lexikographischen Werkes ist der/die Lexikograph/in (oder ein Laie, wie bei einigen kollaborativen Werken) als das denkende Agens zu verstehen. Was die Benutzung betrifft, ist bei der Lexikographie der primäre Adressat der Handlung ein menschlich denkendes Wesen, der sekundäre Adressat eine Maschine (s. Abbildung 2). Im Falle der KI konzipiert ein/e Entwickler/in von Computersystemen und Software ein intelligentes System. Bei der KI ist unseres Erachtens der primäre Adressat ein nicht denkendes Wesen — eine Maschine —, sekundäre Adressaten sind wieder Maschinen, aber auch Menschen. Unangesprochen agieren die primären und sekundären Adressaten in Hinblick auf die Denkprozesse anders (wenn die Rede überhaupt in allen Fällen von Denkprozessen sein kann).

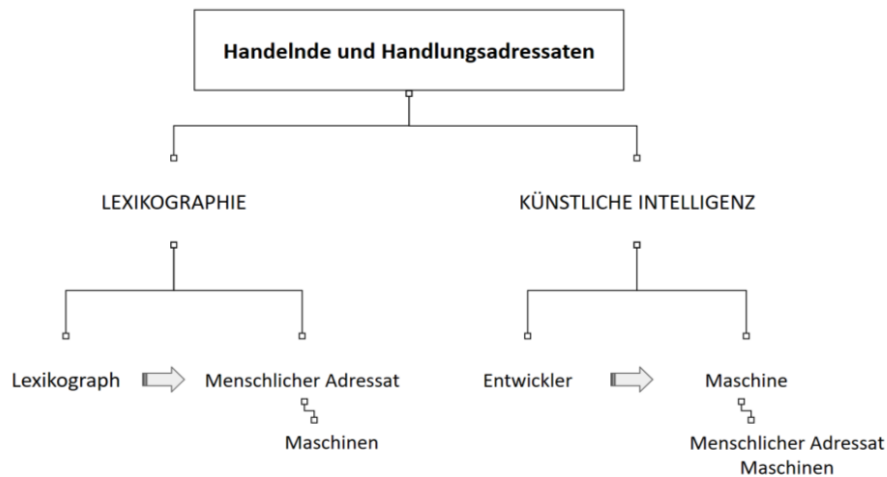

**Abbildung 2:**    Handelnde und Adressaten in der Lexikographie und der KI

Zieht man die Unterscheidung zwischen Computerlexikographie und computergestützten Lexikographie in Betracht, ist das vorangehende Schema wie folgt weiter zu ergänzen:

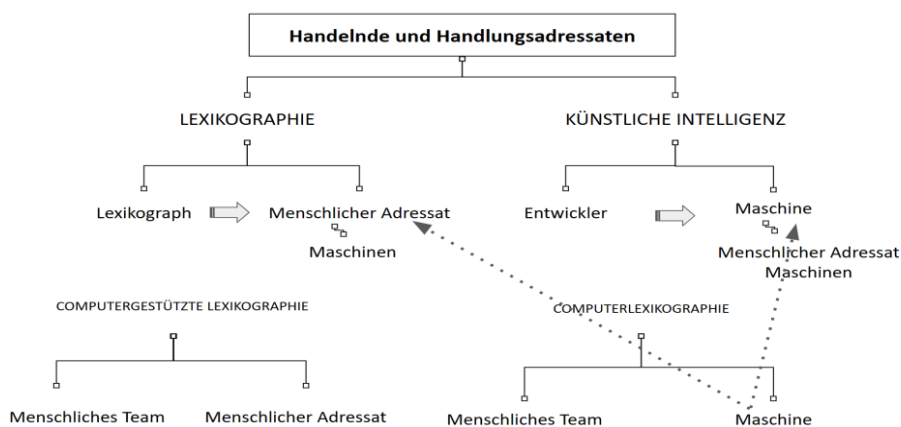

**Abbildung 3:** Handelnde und Adressaten in der Computerlexikographie, der computergestützten Lexikographie und der KI

Sowohl LexikographInnen[18] als auch menschliche BenutzerInnen[19] von lexikographischen Ressourcen möchten Antwort auf eine Frage finden. In diesem Zusammenhang stellen das lexikographische Team sowie die menschlichen BenutzerInnen Hypothesen auf. Im Falle der KI wird die einzige kognitive Hypothesenbildung vom Entwicklungsteam und von sekundären menschlichen BenutzerInnen hergestellt. Maschinen per se bilden keine Hypothese: Intelligente Systeme verarbeiten auf der Grundlage von Algorithmen Daten und in Anlehnung daran bieten sie ein Output. Kreativ — im Sinne von kognitiven Prozessen — sind sie aber nicht.

Dies lässt sich bei dem Output in den ChatGPT-Sessions (s. 4.3 und 4.4) beobachten. Das System aktiviert bei der Beantwortung des Prompts einen konkreten Wissensbereich. Wenn nicht genügende Informationen vorhanden sind, dann wendet ChatGPT Ähnlichkeitsprinzipien an; das erklärt, warum es in den galicischen Antworten auf das Portugiesische oder auf das Spanische zurückgreift. Diese Ergebnisse lassen sich angesichts der Angemessenheit hinsichtlich der Kohärenz, Kohäsion oder Textsortenspezifizität nur von einem menschlichen Agens evaluieren[20] (vgl. Effizienz in 5.4). Diese kognitive Evaluationshandlung erlaubt den Schluss, dass die von ChatGPT generierte Antwort im Galicischen hinsichtlich „Wörterbuchartikel" auf „Artikel" im Sinne von ‚Abhandlung, Aufsatz' zurückgeht (vgl. 4.4). Dies erklärt die ChatGPT-Vermittlung der Informationen in vollständigen Absätzen in wissenschaftlicher

Form, was der lexikographischen Tradition eher widerspricht: In Wörterbuchartikeln wird jedem Textteil eine Funktion zugewiesen und die Information wird meist kondensiert.

### 5.2    Der genuine Zweck und der Intelligenzbegriff

Zur Auseinandersetzung mit dem Intelligenzbegriff sei an den genuinen Zweck eines Wörterbuches erinnert:

> Der genuine Zweck eines Wörterbuches besteht darin, dass es benutzt wird, um anhand lexikographischer Daten in den Teiltexten mit äußerer Zugriffsstruktur […] Informationen zu denjenigen Eigenschaftsausprägungen bei sprachlichen Ausdrücken zu erschließen, die zum jeweiligen Wörterbuchgegenstand gehören. (Wiegand 1998: 299)

Überträgt man diese Definition auf intelligente Systeme, kommt man zur Schlussfolgerung, dass der „genuine Zweck" eines intelligenten Systems darin besteht, anhand einer großen Datenmenge und nach einer Phase maschinellen Lernens Handlungen teilweise auszuführen, die Menschen bei unterschiedlichen Aktivitäten unterstützen können. Derartige Systeme haben aber keineswegs die Absicht, lexikographisches Wissen zu vermitteln, und ihre Benutzung als Nachschlagewerke gewährleistet nicht, dass die Benutzenden Einsichten in das sprachliche Verhalten einer gewissen lexikalischen Einheit oder in die Wörterbuchartikel als Textsorte gewinnen. Das lässt sich bei der Analyse der Ergebnisse in vorangehenden Abschnitten beobachten.

### 5.3    Daten: Präsentation, Systematisierung und Qualität

Die Pilotexperimente zeigen, dass eine Systematisierung im Output von ChatGPT fehlt und dass das System auf wenigen lexikographischen Texten trainiert wurde. Die Metriken (s. 4.2) weisen eine geringe Ähnlichkeit zwischen den Outputs in den zwei Sitzungen auf, die sich auch bei der qualitativen Analyse nachweisen lässt. Hinsichtlich dieser mangelnden Einheitlichkeit vertreten Jakubiček und Rundell (2023) die Ansicht, dass ChatGPT keine festgelegten Kriterien hat. Das lässt sich z.B. an den generierten Daten der mikrostrukturellen Textteile beobachten, indem die Lemmazeichengestaltangabe und die Bedeutungsangabe die einzigen regelmäßig vorhandenen Angaben sind. Es lässt sich vermuten, dass diese die einzigen beständigen lexikographischen Textteile in den Trainingsdaten vom ChatGPT darstellen. Dennoch sind die Output-Unterschiede in den beiden analysierten Sprachen nicht zu übersehen (vgl. 4.3 und 4.4). Dies geht unseres Erachtens darauf zurück, dass die meisten Daten im Chatbot auf Englisch sind (vgl. Jakubiček und Rundell 2023), dass das System *crosslingual* operiert und dass es auf wenigen Daten auf Galicisch beruht. Bezüglich der Datenqualität sei hier zu betonen, dass die Mehrheit der identifizierten Fehler auf der semantischen Ebene zu verorten ist.

**5.4    Effizienzgrad und Lösungsverfahren**

Die wenig systematisierten und begrenzt einheitlichen Outputs machen deutlich, dass das Verfahren zur Lösung lexikographischer Probleme bzw. Fragen bei ChatGPT nicht effizient ist (vgl. Radford et al. 2018): Je nach Kontext und bedingt durch das vorherige Gespräch mit den BenutzerInnen antwortet ChatGPT anders auf die Fragen. Das bedeutet dann, dass der Effizienzgrad intelligenter Systeme nicht nur nach der Datenmenge evaluiert werden kann (vgl. Phoodai und Rikk 2023).

Zur endgültigen Evaluation des Effizienzgrades muss außerdem die Rolle der Benutzenden in Erwägung gezogen werden. Benutzende/Handelnde agieren nicht passiv, denn sie wollen sich neue Informationen erschließen. Die denkenden Handelnden — sowohl LexikographInnen als auch BenutzerInnen — können die angebotenen Informationen auswerten und der Handlung entsprechend agieren. Unterscheidet man zwischen Laien und ExpertInnen[21], wird der Effizienzgrad des Problemlösungsverfahrens davon abhängen, wie kundig ein bestimmter Benutzer oder eine bestimmte Benutzerin ist. Folglich kann ein/e erfahrene/r Benutzer/in imstande sein, sich die Informationen zu erschließen, die vom intelligenten System geliefert werden. Hingegen kann ein/e unkundige/r Benutzer oder Benutzerin — zum Beispiel ein/e Sprachlernende/r mit begrenzter Sprachkompetenz– nicht validieren, ob die Informationen den Tatsachen entsprechen oder nicht.

**6.    Zusammenfassung**

Neben der quantitativen und qualitativen Evaluation befasst sich der Aufsatz auch mit Gemeinsamkeiten und Unterschieden zwischen der Lexikographie und der künstlichen Intelligenz, somit werden nicht nur der Effizienzbegriff im Sinne des Wirkungsgrades und der Dateninterpretation, sondern auch der Intelligenzbegriff, die Handlung und die Handelnden bei kognitiven Prozessen diskutiert.

Die Analyse von ChatGPT als hilfeleistendem Werkzeug bei einem lexikographischen Projekt ist nicht das Ziel der Untersuchung. Aus diesem Grund ist das Chatbot mit unterschiedlichen Prompts nicht traniert worden. Offene Fragen für künftige Studien bestehen in der Interaktion zwischen der Lexikographie und ChatGPT, z.B. inwiefern die Lexikographie zwecks der Überarbeitung und Anpassung generierter Wörterbuchartikel von diesem Chatbot profitieren kann. Es ist festgestellt worden, dass ChatGPT über Daten bezüglich der Textsorte „Wörterbuchartikel" im Deutschen und im Galicischen nicht verfügt. Die Datenverfügbarkeit bei anderen Sprachen bedarf einer weiteren Untersuchung.

Insgesamt lässt sich festhalten, dass zurzeit ChatGPT die lexikographische

Arbeit nicht ersetzen kann: ChatGPT kann nicht als Garant für das kulturelle und sprachliche Kulturerbe gelten, es gilt nicht als ein kultureller, sozialer, normativ-linguistischer und sogar rechtlicher Bezugspunkt.

Diese Studie zeigt, dass heutzutage die Lexikographie von der künstlichen Intelligenz nicht ersetzbar ist, denn die KI kann eine vollständige lexikographische Tätigkeit nicht ausführen. Das ist eben nicht ihr Ziel. Beide weisen unterschiedliche Aktoren, Handlungen, Forschungsgegenstände, Analyseverfahren, Ziele und Ergebnisse auf.

## Danksagung

## Endnoten

1. ChatGPT ist ein von OpenAI entwickelter Chatbot, der Antworten in natürlichen Sprachen generieren kann. In dieser Studie arbeiten wir lediglich mit der kostenfreien Version 3.5.
2. Im Bereich der KI und der natürlichen Sprachverarbeitung spricht man von Prompt, um eine Eingabeaufforderung zu benennen, die ein Mensch an ein System — in diesem Fall an ChatGPT — richtet und die den Chatbot dazu veranlasst, eine Antwort zu generieren.
3. Weitere Beispiele sind Kabashi (2018) und Delli Bovi und Navigli (2017).
4. Eine Ausnahme bilden Müller-Spitzer et al. (2018). Sie beziehen sich aber auf die Verwendung von online Sprachressourcen, nicht konkret auf ChatGPT.
5. Im WLWF-4 (2020: 120) wird unter „Wörterbuchartikel" das Folgende verstanden: „aus der geordneten Menge von Textelementen und/oder Textbausteinen bestehender Textteil des Wörterverzeichnisses, für den das Lemma obligatorisch ist und in dem mindestens eine Eigenschaft des Lemmazeichens beschrieben oder angeführt wird."
6. *n*-Gramme sind aufeinanderfolgende Wortsequenzen, wobei *n* die Fenstergröße der Sequenzen kodiert. Es bestehen bereits lexikalische Ressourcen und Systeme, die linguistische Daten auf der Grundlage von *n*-Grammen verarbeiten und darstellen (vgl. Wolfer et al. 2023).
7. Diese Metriken werden auch zur Analyse der Ähnlichkeit von zwei Texten herangezogen (vgl. Stefanovič et al. 2019).
8. Auf Anfrage stellen wir den R-Code zur Berechnung der Ähnlichkeitskoeffizienten *Jaccard*, *Dice* und *ROUGE-N* zur Verfügung.
9. Diese Methode wurde auch in anderen Studien lexikographischer Art (vgl. Phoodai und Rikk 2023) angewandt.

10. Für unsere Belange muss geprüft werden, in welchem Ausmaß eine Ähnlichkeitstendenz zwischen den gesamten Wörterbuchartikeln festzustellen ist. Nimmt man spezifische Textteile (wie etwa die Definition), könnte das Verhalten auf einen höheren Grad von Ähnlichkeit verweisen.
11. Die zugrundeliegenden ChatGPT-Sessions sind unter folgenden Links abrufbar:
    (a) Session 1: https://chat.openai.com/share/9c89990d-7e65-4173-9578-d8e613b3a880
    (b) Session 2: https://chat.openai.com/share/c087e4f4-2069-4605-aead-268c9ea1cd47.
12. Es ist hier erwähnenswert, dass im Beispiel nicht einmal das beschriebene Lemma vorkommt.
13. Die ChatGPT-Sessions mit Galicisch als Metasprache sind unter folgenden Links abrufbar:
    (a) Session 1: https://chat.openai.com/share/0ec3c864-0508-4cef-b8d5-c93c43e05c3e,
    (b) Session 2: https://chat.openai.com/share/55313ca3-90dc-4fcf-9dbb-36721b03ee7e.
14. Übersetzung von uns: „Das Wort L ist eines der mächtigsten und häufigsten Wörter in der galicischen Sprache. Da es so häufig verwendet wird, ist es wichtig, seine vielfältigen Bedeutungen und Verwendungen zu verstehen. In diesem lexikographischen Artikel werden wir das Wort L im Galicischen untersuchen und seine vielfältigen Bedeutungen, grammatikalischen Konstruktionen und Verwendungsbeispiele analysieren."
15. Übersetzung von uns: „Das Verb L ist aufgrund seiner Vielseitigkeit und Bedeutung in der Kommunikation ein zentrales Element der galicischen Sprache. Seine Konjugation und sein vielfältiger Gebrauch machen es zu einem unverzichtbaren Werkzeug, um eine Vielzahl von Handlungen, Ursachen und Zuständen auszudrücken. Es ist ein Wort, das über die Grammatik hinausgeht und für die Sprecher des Galicischen zu einem festen Bestandteil des täglichen Lebens wird, was zum Reichtum und zur Vielfalt dieser Sprache beiträgt."
16. Übersetzung von uns: „Es ist eines der gebräuchlichsten und wichtigsten Verben in der galicischen Sprache und spielt eine grundlegende Rolle in der Kommunikation und der Übermittlung von Informationen."
17. Damit ist gemeint, dass das ChatGPT-System in jeder Sitzung auf einen anderen Datenabschnitt zurückgreift, sodass die bereitgestellten Informationen von vorherigen abweichen (können).
18. Im Falle der LexikographInnen geht es darum, wie bestimmte Inhalte lexikographisch vermittelt werden können, wer der Adressat des Werkes sein wird, u.a. Dabei versucht man neben computerbezogenen Fragen im lexikographischen Prozess (Müller-Spitzer 2007: 17), auch weitere, „die darauf abzielen, diejenigen theoretischen, methodischen, terminologischen, historischen, dokumentarischen, didaktischen und kulturpädagogischen Fragen zu beantworten" (Müller-Spitzer 2007: 17).
19. Die BenutzerInnen machen unterschiedliche kognitive Schritte bei einer Suchanfrage wie z.B.: „Was möchte ich konkret nachschlagen?", „Wie gehe ich bei einer erfolglosen Suche vor?", u.a. Dabei stellen sie eine Hypothese oder Teilhypothesen im ganzen Verfahren auf (Domínguez Vázquez und Valcárcel Riveiro 2015; Müller-Spitzer et al. 2018).
20. Führt man dennoch die Suchanfrage in einem intelligenten System aus, wird man damit konfrontiert, dass man nicht nur Informationen aus den Daten extrahieren sollte, sondern man muss auch bewerten, inwiefern die gelieferte Information richtig ist. Benutzende sollten sich dessen bewusst sein, dass die Informationen, die sie erhalten, nicht unbedingt richtig sein müssen und dass sie zur Formulierung einer fundierten Antwort andere Ressourcen konsultieren sollten.

21. Unter die Gruppe der Laien fallen beispielsweise Sprachlernende oder SchülerInnen. Zur Gruppe der ExpertInnen gehören LexikographInnen, SpezialistInnen und zum Teil auch erfahrene BenutzerInnen.

## Literaturverzeichnis